# A Convolutional Neural Network Approach Towards Self-Driving Cars

Akhil Agnihotri
Department of Mechanical Engineering
Birla Institute of Technology and Science – Pilani
Hyderabad, India
agnihotri.akhil@gmail.com

Prathamesh Saraf
Department of Electrical & Electronics Engineering
Birla Institute of Technology and Science – Pilani
Hyderabad, India
pratha1999@gmail.com

Kriti Rajesh Bapnad
Department of Electrical & Electronics Engineering
Birla Institute of Technology and Science – Pilani
Hyderabad, India
bapnadk@gmail.com

*Abstract*—A convolutional neural network (CNN) approach is used to implement a level 2 autonomous vehicle by mapping pixels from the camera input to the steering commands. The network automatically learns the maximum variable features from the camera input, hence requires minimal human intervention. Given realistic frames as input, the driving policy trained on the dataset by NVIDIA and Udacity can adapt to real-world driving in a controlled environment. The CNN is tested on the CARLA open-source driving simulator. Details of a beta-testing platform are also presented, which consists of an ultrasonic sensor for obstacle detection and an RGBD camera for real-time position monitoring at 10Hz. Arduino Mega and Raspberry Pi are used for motor control and processing respectively to output the steering angle, which is converted to angular velocity for steering.

## I. Introduction

As of today, the performance of CNNs has revolutionized image and pattern recognition by surpassing human performance on standard datasets [1]. The strongest feature of CNNs is that they learn features automatically from training examples and hence override the human need to only select intuitive features for the model [2]. Another benefit of CNNs is that they take advantage of the 2D structure of images and thus have a higher accuracy than a standard flattened neural network. This approach works better than the explicit feature decomposition approaches such as detection of lanes or neighboring cars, since the network decides the most optimal features to extract from the image for itself.

This technology provides a robot with senses with which it can traverse in unfamiliar surroundings without damaging itself. This work contains the details of the developed CNN, design of the robot and an experiment proposal to test the autonomy of the robot in a controlled real environment. The autonomous vehicle developed in this project is expected to satisfy the following objectives:

1. To have the capacity to detect obstacles along with the speed of obstacle in its path based on a predetermined threshold distance.
2. After obstacle and speed detection, changing its course to a relatively open lane on the road by making a decision autonomously.
3. Requiring no human intervention during its operation.
4. Measuring the distance between itself and the surrounding objects in real-time.

## II. System Architecture

### A. Overview

The dataset given on Udacity's GitHub repository is used for the CNN with 80% used for training and 20% for validation [3]. The training data, in addition to the data from the human driver, consists of images of the vehicle car in various shifts from the center of the lane and rotations from the direction of the road. Time-stamped steering angles are extracted from the .bag files of the training data using Robotic Operating System (ROS) and paired with the corresponding image to form a tuple of (image, steering angle) for input to the CNN, which then computes a proposed steering command.

The system is built using the following (see *Fig. 1 and 2*):

1. Circular metal chassis of diameter 25cm.
2. Two Stepper motors (Nema 17) for the wheels.
3. Two Servo motors attached to the stepper motors to give the steering angle to the vehicle.
4. Ultrasonic sensor HC-SR04, mounted on a servo motor on the chassis to check for the obstacles.
5. Web camera for image capturing of lanes on the road and Raspberry-pi 3B for image processing of the images captured and getting the steering angle.

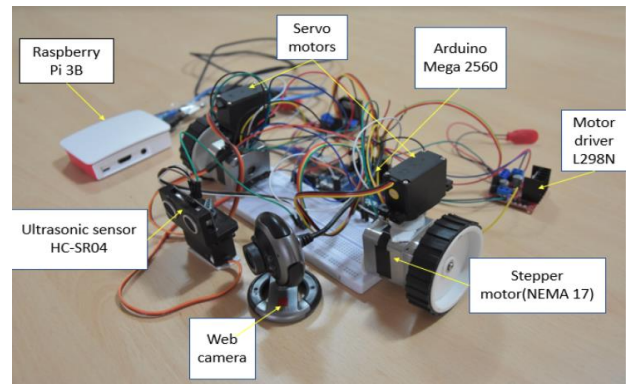

**Fig 1**: Components of the robot

### B. Details of Operation

The network is trained to minimize the mean squared difference of the actual and predicted steering angle as shown in *Fig. 2* [2].

The developed CNN architecture consists of four convolution layers and two fully-connected vanilla layers, each having a different bias, weight and Rectified Linear Unit (ReLU) associated with it. Initially, independent and identically distributed (i.i.d.) weights and bias of 0.1 are assigned. The first three layers are padded with 5x5 kernels with a stride of 2 while the next layer has 3x3 kernels with a stride of 1. These layers are followed by two fully-connected layers which are non-strided to prevent overfitting on the training data, which is segmented into batches of size 100 and trained for 30 epochs on the CNN. These fully connected layers also serve as a controller for the steering, thus eliminating the use of an explicitly defined controller, normally a Constrained Linear Quadratic Regulation (LQR) controller [4]. *Fig 3.* shows the detailed structure of the CNN.

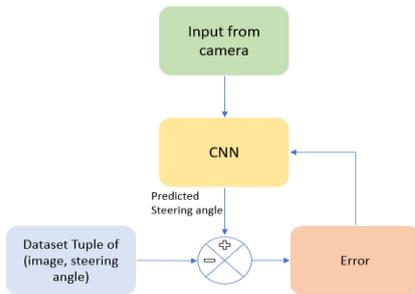

**Fig. 2**: Overview of the CNN model

Three servo motors, two stepper motors and three ultrasonic sensors are connected to the Arduino Mega 2560 board [5]. The ultrasonic sensor is mounted on the servo motor and hence is free to move right and left to check for free space around the vehicle. Sensor readings and steering angle after image processing by the Raspberry-pi and output of the CNN are fed to the servo motors and the stepper motors by Arduino.

The decision tree behind the code is shown in *Fig. 4* and goes as follows:

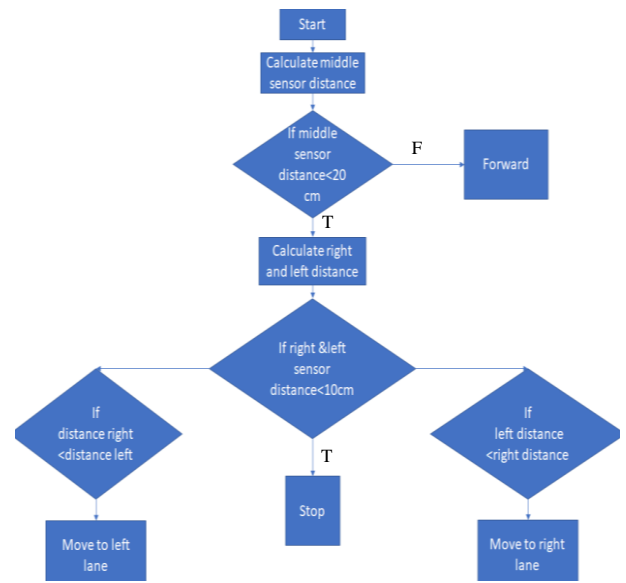

**Fig. 4**: Decision flow of the sensor input and output. 'T' equals True and 'F' equals False in the figure.

The ultrasonic sensor keeps track of the distance between the car and a possible obstacle in its current lane. For forward movement, the servos on top of the stepper motors point at 90° with the motors running at 60 RPM. If the robot encounters a stationary obstacle (based on the constant rate of decreasing distance), the ultrasonic sensor checks for an obstacle-free left or right lane for a distance of 20cm and moves to the lane which is obstacle-free at an angle of 60° to the direction of motion. Secondly, if the robot encounters a moving obstacle within 20cm of its current lane, it is programmed to match the speed of that obstacle and continue in its current lane (see *Fig. 5)*.

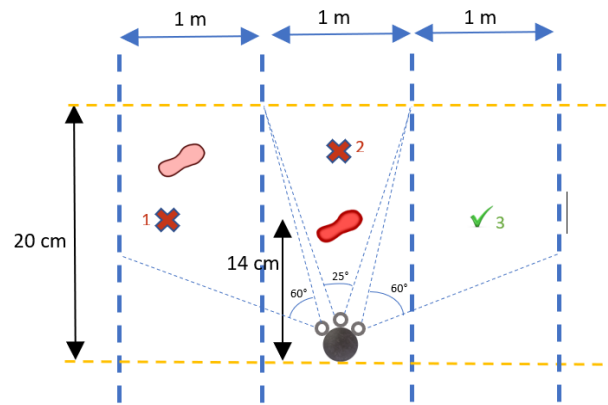

**Fig. 5**: Schematic showing the robot as a grey disk (bottom center) with three ultrasonic sensors attached to its periphery which divide the line of sight into three zones as shown. In this scenario, there is an obstacle at a distance of 14cm in the current lane, along with another obstacle within 20cm of its current horizontal location. This results in publishing of negative angular velocity of .5 radians per second towards the right after which the CNN takes over to move the robot into lane 3. All values are in SI units.

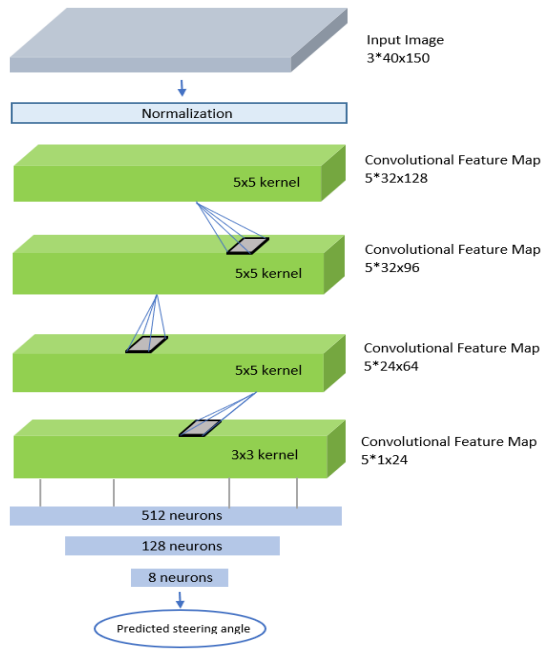

**Fig. 3**: Architecture of the CNN containing 133,120 weights and about 11 million parameters. In the figure, A*BxC indicates A layers of B width and C length each.

## III. RESULTS AND DISCUSSION

The CNN was trained as described above and tested on a dataset provided by the University of Cambridge to give a sub-par performance since the weather conditions of the two datasets are different [6]. The model was trained on two datasets from Udacity [3] and NVIDIA [7]. The Udacity dataset consisted of 6.6 Gigabytes of 33,478 images while the NVIDIA dataset had 2.1 Gigabytes of 7,064 images. As seen in *Fig. 6*, there is a stark difference in the validation Mean Square Error (MSE) loss of the model after 30 epochs in both the instances visualized with the help of TensorFlow [8]. Udacity dataset gave a loss of 0.0003798 after 30 epochs while NVIDIA dataset gave a loss of 1.013 which was approximately 5 times smaller than the Udacity dataset.

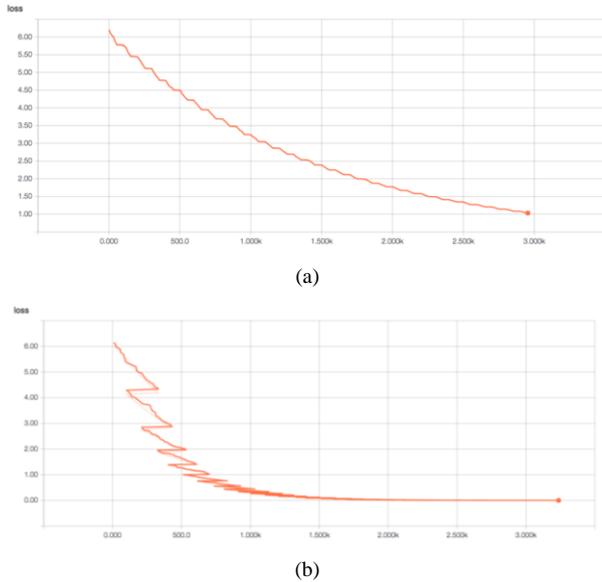

(a)

(b)

**Fig. 6**: MSE over validation data is plotted on the vertical axis and visualized in Tensor Board with number of iterations on the horizontal axis using (a) NVIDIA (b) Udacity dataset. Aside from a lower MSE after an equal number of epochs, the Udacity validation dataset has a larger variance, hence more information, as seen by the erratic fluctuation of MSE in (b).

*Fig. 7* shows a still from the CARLA simulator in which the CNN was tested and an autonomy (1) of 86.43% was obtained based on the empirical result that human intervention in the simulator was required for five seconds on average.

$$autonomy = \left(1 - \frac{number\ of\ interventions\ .\ 5\ seconds}{total\ elapsed\ time\ (seconds)}\right).100 \quad (1)$$

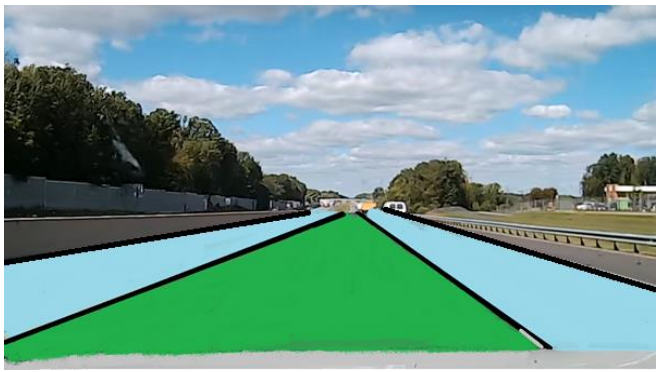

**Fig. 7**: Screenshot of the CNN being implemented in CARLA which is able to steer along the curvature of the roads. The maximum speed in the simulator was set at 30 kilometer per hour and object detection distance was kept at 20 meters. The lanes are bounded by black lines and current safe lane is shown in green while the other two lanes are shown in blue.

## IV. CONCLUSION AND FUTURE WORK

This paper describes a methodology of implementing a level-2 autonomous vehicle in a relatively sparsely occupied environment. A CNN is trained on a dataset by Udacity and used to compute the optimal steering angle based on the image input through the camera. In case of obstruction in the path, three ultrasonic sensors are used to decide in which direction the vehicle should turn to continue on its path. Once this is achieved, the vehicle resumes its normal functioning of manoeuvring based on the steering angle given by the CNN.

Based on the above results, an experiment is being designed for on-road testing on the intra-campus roads of BITS Pilani Hyderabad Campus. The test-path is 580 meters in length with negligible elevation gradient since the dataset was also created on a plane terrain. The path will be marked with three parallel lanes of width 1 meter of Snowcal powder. The GPS module GSM SIM908 will be used for location and path-tracking of the vehicle.

Further research to extend the current CNN to region-based CNNs for posture-recognition passenger safety is also being done. More work is needed to improve the robustness of the network, and to better visualize how the CNN operates internally. Moreover, including other vehicle dynamics parameters in the dataset such as acceleration and velocity will provide for navigation based on real-life constraints on the vehicle.

ACKNOWLEDGMENT

The authors would like to sincerely thank Prof. N.L. Bhanumurthy and Prof. Aruna Malapati of the Department of Computer Science and Information Systems, BITS Pilani Hyderabad Campus for their valuable support and guidance throughout the project.